\documentclass[10pt,twocolumn]{article}

\usepackage[margin=0.72in,columnsep=0.23in]{geometry}
\usepackage[utf8]{inputenc}
\usepackage[T1]{fontenc}
\usepackage{amsmath,amssymb}
\usepackage{booktabs}
\usepackage{graphicx}
\usepackage{microtype}
\usepackage{times}
\usepackage[scaled=0.92]{inconsolata}
\usepackage{xcolor}
\usepackage{hyperref}
\usepackage{xurl}
\usepackage[font=small,labelfont=bf]{caption}
\usepackage{enumitem}
\usepackage{placeins}
\usepackage{listings}

\definecolor{codeblue}{HTML}{205493}
\definecolor{codegray}{HTML}{5F6B76}
\definecolor{codeback}{HTML}{F5F6F7}

\lstdefinestyle{kerascode}{
  language=Python,
  basicstyle=\ttfamily\scriptsize,
  keywordstyle=\bfseries\color{codeblue},
  commentstyle=\color{codegray},
  stringstyle=\color{red!45!black},
  columns=fullflexible,
  keepspaces=true,
  showstringspaces=false,
  breaklines=true,
  frame=single,
  framerule=0.4pt,
  rulecolor=\color{black!25},
  backgroundcolor=\color{codeback},
  xleftmargin=2pt,
  xrightmargin=2pt,
  aboveskip=4pt,
  belowskip=4pt
}

\hypersetup{
  colorlinks=true,
  linkcolor=codeblue,
  citecolor=codeblue,
  urlcolor=codeblue,
  pdftitle={Many Optimizers But Only One Training Path: Repeated Resampling for Adaptive Optimizer Selection},
  pdfauthor={Ronald Richman and Mario V. Wuthrich},
  pdfkeywords={optimizer selection, repeated resampling, optimizer state, insurance pricing}
}
\setlist{nosep,leftmargin=*}
\title{\vspace{-1.3em}
Many Optimizers But Only One Training Path:\\
Repeated Resampling for Adaptive Optimizer Selection}
\author{
Ronald Richman\\
\small insureAI
\and
Mario V. W\"uthrich\\
\small ETH Z\"urich
}
\date{August 2026}

\begin{document}
\raggedbottom
\maketitle

\begin{abstract}
An optimizer is usually chosen before training a deep neural network and then kept fixed. Treating optimizer
choice as a hyperparameter could boost performance, but it requires several complete
training runs and discards all but the winner. Repeated Optimizer Resampling
(ROR) instead searches during one evolving run. Every \(b\) epochs, each
candidate optimizer scouts from the current model weights for \(s\) epochs.
The best scout continues for the remaining \(b-s\) epochs, and that completed
segment becomes the new incumbent if it improves the validation objective.
This design allows the preferred optimizer to change as training progresses.

We compare two variants of ROR on MNIST, Fashion-MNIST, and two
motor insurance claim-count models. Nine fixed optimizers and both ROR variants are evaluated with the same ten seeds. 
One-epoch ROR uses 24\% to 35\% of the aggregate training needed
to identify the best fixed optimizer exhaustively and remains close to that
optimizer on all four tasks. These results support short
scouting as a practical way to search over optimizers without completing every
candidate run.
\end{abstract}

\section{Introduction}

Training deep neural networks requires choosing an optimizer. In practice,
that choice is often made heuristically, even though it is an important
hyperparameter, and the selected optimizer is usually kept fixed for the whole
training run. AdamW has emerged as a standard choice
\cite{loshchilov2017decoupled}, while recent work reports strong results for
Muon when training Transformer models
\cite{liu2025muon,jordan2024moddednanogpt}. Some of the choices of optimizers that are available
include Adam \cite{kingma2014adam}, Lion \cite{chen2023lion}, momentum SGD,
Nadam \cite{dozat2016nadam}, RMSprop \cite{hinton2012rmsprop}, LAMB
\cite{you2019lamb}, and ScheduleFreeAdamW
\cite{defazio2024schedulefree}. These methods use different gradient
transformations, update rules, and forms of optimizer state.

The best choice for a new problem is not known in advance. It can be identified
retrospectively by training every candidate to convergence, but this
conventional hyperparameter search is expensive because all but the winning
run are discarded. The optimizer with the best complete run also need not be
best throughout training: early descent and late refinement may benefit from
different update rules. An adaptive sequence could therefore outperform every
fixed optimizer, or at least approach the best fixed result without first
completing the entire search.

To test this idea, we adapt a procedure from general artifact optimization
\cite{agrawal2026optimizeanything}. Repeated optimizer resampling (ROR) opens a
short tournament at regular intervals. Each candidate optimizer scouts from
the same model weights. The best scout continues to the end of the next
training segment, while the other scouts are discarded. Repeating the
tournament produces one model trajectory and a data-dependent optimizer
schedule. ROR is structurally close to successive halving and Hyperband
\cite{li2017hyperband}, which also train many candidates on short budgets and
promote only the most promising ones to longer training budgets. The difference is ROR's repeated re-entry of optimizer candidates into subsequent tournaments:
Hyperband eliminates a losing configuration permanently, whereas every
optimizer re-enters each ROR tournament from the current incumbent weights,
so a candidate that loses early can still be selected later in training.

An optimizer consists of more than its update formula. Momentum buffers,
adaptive moment estimates, iteration counters, and schedule-free variables
store information from earlier gradients. If the same optimizer wins two
consecutive ROR rounds, resetting these variables makes the second branch a
fresh optimization run from the retained weights. Preserving them instead
continues the optimization process begun in the preceding round. We therefore test two state policies in what follows. State-preserving ROR
(SP-ROR) retains the incumbent optimizer's compatible state when it is selected
again. Cold-start ROR (CS-ROR) reinitializes every optimizer at the start of
every tournament.

Stochastic gradient descent with warm restarts (SGDR) \cite{loshchilov2017sgdr} provides a useful point
of comparison to ROR. SGDR keeps the current model weights
and optimizer, returns the learning rate to the top of a cosine schedule at a
scheduled restart, and anneals it again. These restarts often lead to improved model performance. ROR is different: a tournament
compares several optimizer families, not several learning-rate phases of one
optimizer, but we expect the model performance to improve, similar to SGDR. 

We ask three questions. Can ROR approach the best fixed optimizer found in
hindsight while using a smaller training budget than the exhaustive search needed to
identify the best fixed optimizer? How much can the scouting period be shortened before predictive
performance changes? Does preserving the incumbent optimizer's state affect
the resulting model or optimizer schedule?

We evaluate nine optimizer and learning-rate pairs on MNIST, Fashion-MNIST,
and two motor insurance claim-count models. The best fixed optimizer is
an oracle benchmark: it is known only after the exhaustive comparison has
finished, and its search cost is the sum of all nine fixed runs. In addition to the oracle benchmark, we will run ROR and, also, a
one-shot baseline that compares the set of optimizers once near the start of training.

\section{Repeated optimizer resampling}

An ROR tournament has two time scales. The \emph{scouting period} \(s\) is the
number of epochs used to compare the candidate optimizers. The \emph{retained
segment length} \(b\) is the number of epochs by which an accepted tournament
advances the model trajectory, with \(1\leq s\leq b\). At the start of a
tournament, ROR copies the current weights \(K\) times. Every optimizer trains
one copy for \(s\) epochs using the same data order. The best scout then
continues for \(b-s\) more epochs. If the completed segment improves the
incumbent validation objective sufficiently, its weights and optimizer state
become the new incumbent. Otherwise, training stops.

Formally, let \(\mathcal{O}=\{o_1,\ldots,o_K\}\) be the candidate set and let
\(J_{\mathrm{val}}\) be oriented so that smaller values are better. At the
start of tournament \(r\), the incumbent has weights \(\theta_r\), optimizer
identity \(o_r\), and optimizer state \(u_r\). The first tournament starts
from the common random initialization \(\theta_0\), without an incumbent
optimizer. Candidate \(o\) receives the state \(\widetilde u_{r,o}\) specified
by the state policy (see the next section) and scouts for \(s\) epochs:
\begin{equation}
  (\theta^{(s)}_{r,o},u^{(s)}_{r,o}) =
  \operatorname{Train}(\theta_r,o,\widetilde u_{r,o},s),
  \qquad o\in\mathcal{O}.
\end{equation}
The scout winner is
\begin{equation}
  o_r^\star = \arg\min_{o\in\mathcal{O}}
  J_{\mathrm{val}}(\theta^{(s)}_{r,o}).
\end{equation}
It continues directly from its scout weights and scout state:
\begin{equation}
  (\theta^{(b)}_{r,o_r^\star},u^{(b)}_{r,o_r^\star}) =
  \operatorname{Train}\!\left(
    \theta^{(s)}_{r,o_r^\star},
    o_r^\star,
    u^{(s)}_{r,o_r^\star},
    b-s
  \right).
\end{equation}
When \(s=b\), this continuation has length zero and the scout itself is the
completed segment. If the completed segment improves the incumbent objective
by more than \(\delta\), ROR commits it:
\begin{equation}
  (\theta_{r+1},o_{r+1},u_{r+1}) =
  (\theta^{(b)}_{r,o_r^\star},o_r^\star,u^{(b)}_{r,o_r^\star}).
\end{equation}
Otherwise, the round is rejected: the incumbent weights and state remain
unchanged, and training stops. The accepted segments define the optimizer
schedule along the retained trajectory.

We measure search cost in epoch-equivalents.\footnote{A fixed number of
gradient-update steps could also define the scouting and segment budgets. We use epochs here
because the experiments and their early-stopping rules are specified in
epochs; fixed-step tournaments are not evaluated.} One tournament trains all
\(K\) scouts for \(s\) epochs and the winner for another \(b-s\) epochs, so
\begin{equation}
  C_{\mathrm{round}}=Ks+(b-s)=b+(K-1)s.
\end{equation}
With nine optimizers and \(b=3\), the cost is 11 epoch-equivalents for
\(s=1\), 19 for \(s=2\), and 27 for \(s=3\). If ROR opens \(T\)
tournaments, its cost is \(T C_{\mathrm{round}}\). By comparison, an
exhaustive optimizer search costs \(\sum_o E_o\), where \(E_o\) is the number
of epochs needed to train fixed optimizer \(o\) to its stopping point. ROR can
therefore use less training than the exhaustive search even though every
optimizer participates in every tournament. Aggregate cost includes accepted
and rejected tournaments.

\paragraph{State-Preserving and Cold-Start ROR.}
The policies differ only in the state supplied at the start of each scout. In State-Preserving ROR (SP-ROR), a scout using the incumbent optimizer resumes from its accumulated
state, while every challenger begins with newly initialized state:
\begin{equation}
  \widetilde u_{r,o}^{\mathrm{SP}} =
  \begin{cases}
    u_r, & o=o_r,\\
    u_o^0(\theta_r), & o\ne o_r.
  \end{cases}
\end{equation}
In Cold-Start ROR (CS-ROR), every branch starts with newly initialized optimizer state:
\begin{equation}
  \widetilde u_{r,o}^{\mathrm{CS}} = u_o^0(\theta_r),
  \qquad o\in\mathcal{O}.
\end{equation}
All states are new in the first tournament. In either policy, the selected
scout keeps its own state during the \(b-s\) continuation. The difference
arises only at the next tournament: SP-ROR can restore the incumbent's
momentum, adaptive moments, iteration count, and schedule variables, whereas
CS-ROR resets them.

We do not retain the state of every losing optimizer. Such state was learned
along discarded weights, so attaching it to the winner's weights would be a
state ``transplant'' rather than exact continuation. Moreover, exact preservation would
require retaining each optimizer's matching weights and advancing all nine
paths in parallel, which would recreate the exhaustive search that ROR is
intended to avoid.

\paragraph{One-shot selection baseline.}
As a baseline, we also test one-shot selection, which queries the optimizer set
only once. Starting from a common random initialization, all $K$ optimizers
train separate model copies for $s$ scout epochs; we set $s=1$, the shortest
scout also used by ROR. The optimizer with the best validation result is selected, its
post-scout model weights are retained, and all other branches are discarded. A
newly initialized instance of the selected optimizer then continues from those
weights until the ordinary early-stopping rule fires. The model is not returned
to its original weights, but the optimizer set is never compared again.

If the selected optimizer trains for $e$ additional epochs, the total cost is
$Ks+e$ epoch-equivalents: $Ks$ scout epochs plus the continuation. Because the
continuation optimizer is newly initialized, one-shot is a low-cost operational
baseline rather than a one-factor ablation of how often the set is queried. The
clean state-policy comparison is SP-ROR versus CS-ROR.

\paragraph{Implementation.}
Every branch starts from the same copied weights. Under SP-ROR, the incumbent
restores its optimizer variables and challengers use new optimizer objects;
under CS-ROR, all optimizer objects are new. Validation selects both the model
and optimizer state for the next round, and the test set is not consulted.
Appendix~\ref{app:pseudocode} gives the core Keras 3 pseudocode.

\section{Experimental design}

We run four experiments in two application settings. MNIST and Fashion-MNIST
provide conventional image classification tasks, with a classification objective. Motor third party
liability claim counts provide a well-studied insurance-pricing problem from
actuarial science, allowing us to examine ROR on structured tabular data using
Poisson deviance and two model architectures.

Each experiment has the same comparison structure. The nine fixed optimizers
are trained independently to define the hindsight-best fixed result and the
aggregate cost of finding it. One-shot selection provides a low-cost reference.
CS-ROR shows what repeated resampling does when every branch restarts its
optimizer, while SP-ROR shows what changes when a repeated incumbent continues
with its state intact.

\paragraph{Image classification data and model.}
MNIST \cite{lecun1998gradient} and Fashion-MNIST
\cite{xiao2017fashion} each contain 60,000 training and 10,000 test
$28\times28$ grayscale images. For each seed, 5,000 training observations are
randomly reserved for validation. The remaining observations are presented in
a fixed order to every branch. The model rescales pixels, flattens each image,
and applies dense GELU layers of widths 256, 256, and 128 before a ten-class
linear output. It is therefore a fully connected multilayer perceptron, not a
convolutional neural network. This deliberately simple architecture lets us
study optimizer selection without convolution-specific design choices. The
model has 300,938 trainable parameters. The model is trained using the cross-entropy loss.

\paragraph{Insurance pricing data and models.}
The cleaned French motor third party liability frequency data
\cite{dutang2024insurance,wuthrichmerz2023} contain 678,007 policies. We use
the published split: 610,206 learning rows and 67,801 test rows. For each seed,
61,021 rows from the learning set are held out for validation, leaving 549,185
for fitting. The model receives Area, vehicle brand, fuel type, and Region as
categorical variables. Vehicle power, vehicle age, driver age, bonus-malus,
and population density are numerical variables. Policy ID and total claim
amount are excluded.

Each categorical variable has its own trainable embedding. A numerical value
is placed between up to 16 training-set quantile knots and represented by a
linear interpolation of the two adjacent trainable embeddings
\cite{gorishniy2022embeddings}. The
4,791-parameter MLP concatenates these feature embeddings and uses hidden
widths 64 and 32. The 4,433-parameter Transformer \cite{vaswani2017attention} maps every
feature to a 16-dimensional token, adds a learned position embedding, and prepends a
learned CLS token. One two-head self-attention block and a feed-forward block
update the ten tokens, and the decoder reads the CLS token. This construction
follows the tabular tokenization in
\cite{richman2025credibility}, without their credibility
sampling mechanism.

For policy $i$ with exposure $v_i$, the network output $f_\theta(x_i)$ is a log
frequency and
\begin{equation}
  \log \mu_i = f_\theta(x_i) + \log v_i .
\end{equation}
Exposure is an offset rather than an ordinary feature. We minimize
$\mu_i-y_i\log\mu_i$ and report mean Poisson deviance on the held-out test set.
Calibration is total predicted claims divided by total observed claims.

\paragraph{Optimizer set.}
We use Adam at $10^{-3}$; AdamW at $10^{-3}$ with weight decay $10^{-4}$;
Lion at $3\times10^{-4}$ with weight decay $10^{-4}$; Muon at $10^{-3}$ with
weight decay $10^{-4}$; and SGD-Nesterov at $5\times10^{-2}$ with momentum
$0.9$. Following the hybrid parameter treatment used for Muon
\cite{liu2025muon}, Muon is applied only to hidden two-dimensional kernels;
its AdamW branch handles biases and the input and output projections. The set also includes
Nadam at $10^{-3}$, RMSprop at $10^{-3}$, LAMB at $10^{-3}$ with weight decay
$10^{-4}$, and ScheduleFreeAdamW at $2.5\times10^{-3}$ with weight decay
$10^{-4}$. ScheduleFreeAdamW is the native
\texttt{keras.optimizers.ScheduleFreeAdamW} class in Keras 3.15.0, which
implements the method of Defazio et al. \cite{defazio2024schedulefree}. No
separate schedule-free package or local implementation is used.

These nine optimizer and learning-rate pairs form the exhaustive optimizer
comparison, one-shot baseline, and ROR set in every experiment. The learning rates are fixed before
the experiments and are not tuned separately for a dataset or model.

\paragraph{Common evaluation setup.}
All results use seeds 1 through 7, 42, 123, and 2026. The design is fully
crossed: every training rule uses the same ten seeds and the same seed-specific
data split, initialization procedure, and fixed data order. A rejected ROR round
commits neither weights nor optimizer state, so both policies stop at their
first rejection. Test performance is evaluated only after validation-based
selection. We report the mean and sample standard deviation over seeds.
Confidence intervals use paired seed-level differences and the $t_9$
quantile.\footnote{For each comparison, we form the ten seed-matched
differences, compute their sample mean $\bar d$ and sample standard deviation
$s_d$, and report
$\bar d \mathbin{\pm} t_{0.975,9}s_d/\sqrt{10}$.}
The epoch-equivalent totals include every branch trained by the program,
including the final rejected round.

\paragraph{Image classification setup.}
The image experiments use batch size 256, Keras 3.15, and the JAX CPU backend.
Fixed baselines stop when validation accuracy fails to improve by 0.02
percentage points for five epochs, with a 30-epoch cap. The one-shot baseline
uses the single handoff defined above. ROR uses \(b=3\), scouting periods
\(s\in\{1,2,3\}\), \(\delta=0.02\) percentage points, and a maximum committed
trajectory of 30 epochs.

\paragraph{Insurance pricing setup.}
Gorishniy et al. benchmark 15 optimization methods across 17 tabular datasets
and find that Muon consistently outperforms AdamW
\cite{gorishniy2026optimizers}. In our study, both insurance architectures use the
optimizer set described above. Muon receives seven
non-embedding matrices in the attention projections, feed-forward block, and
CLS decoder; embeddings and the final output layer use Muon's AdamW branch.
Batch size is 4,096. Fixed runs stop after five epochs without a validation
loss decrease greater than $10^{-5}$, with a 300-epoch safety cap. Both models
use \(b=3\) and \(s\in\{1,2,3\}\), and ROR accepts a completed segment only if
it lowers the validation loss by more than \(\delta=10^{-5}\), the same
improvement threshold used by the fixed-run stopping rule.

\section{Results}

\subsection{Predictive performance and search cost}

Shorter scouting is the clearest result. With \(b=3\), one-epoch scouting \(s=1\) uses
24\% to 35\% of the training required to complete all nine fixed-optimizer
runs. It also costs 51\% to 68\% less than a ROR design with \(s=3\). A ROR
design with \(s=3\) itself costs less, on average, than completing all nine
fixed-optimizer runs because a ROR training run is on average shorter than a
full fixed-optimizer run, even though each \(s=b=3\) tournament trains all nine
candidates for the full segment. Across the four tasks that average saving is
about 17\% to 42\%; the retained ROR trajectory is typically 8 to 11 epochs on
the image and Insurance Transformer experiments, against about 16 for a typical fixed
run, and about 26 to 28 epochs on the Insurance MLP, against 49. The predictive effect is smaller and depends on the task. Across all four
experiments, no ROR comparison with the best fixed optimizer has a paired 95\%
confidence interval that excludes zero. The same is true when SP-ROR is
compared with CS-ROR. 

The practical implication is that ROR substitutes very effectively for the exhaustive search rather than merely approximating one fixed run: the best fixed optimizer
differs across the four tasks and is only known after all nine runs
complete, while one-epoch ROR achieves similar observed performance, with none of the paired comparisons excluding zero, at 24\% to 35\% of that cost. On the Insurance Transformer, the one-epoch ROR losses are in fact the lowest observed in the table.

The one-shot baseline is cheaper than one-epoch ROR on every task, and the
comparison between the two is informative. On MNIST, one-shot reaches an
observed mean of \(98.116\%\) at 20.3 epoch-equivalents, better and cheaper
than both one-epoch ROR variants. On Fashion-MNIST and both insurance models the
one-epoch ROR means are better, for example a Transformer deviance of
\(0.237627\) against \(0.237968\), at roughly twice the one-shot cost. A
single early tournament therefore captures much of the value of optimizer
selection when the ranking observed after one epoch persists, as the stable
MNIST schedules in Figure~\ref{fig:schedules} suggest. The tasks whose
schedules switch optimizer later in training are the ones where repeated
tournaments improve on the single early query.

\begin{figure*}[t]
\centering
\includegraphics[width=\textwidth]{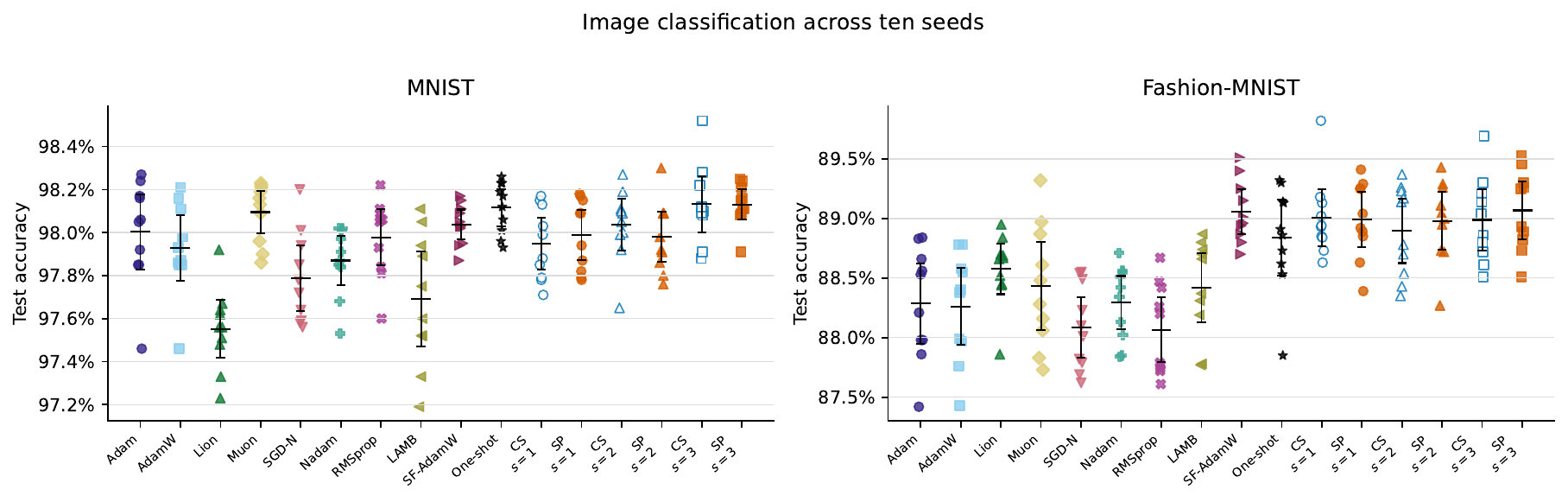}
\caption{Image classification accuracy by seed (bigger is better). Points are individual runs;
horizontal bars show means and 95\% confidence intervals. ROR uses retained
segments of \(b=3\) epochs.}
\label{fig:accuracy}
\end{figure*}

\begin{figure*}[t]
\centering
\includegraphics[width=\textwidth]{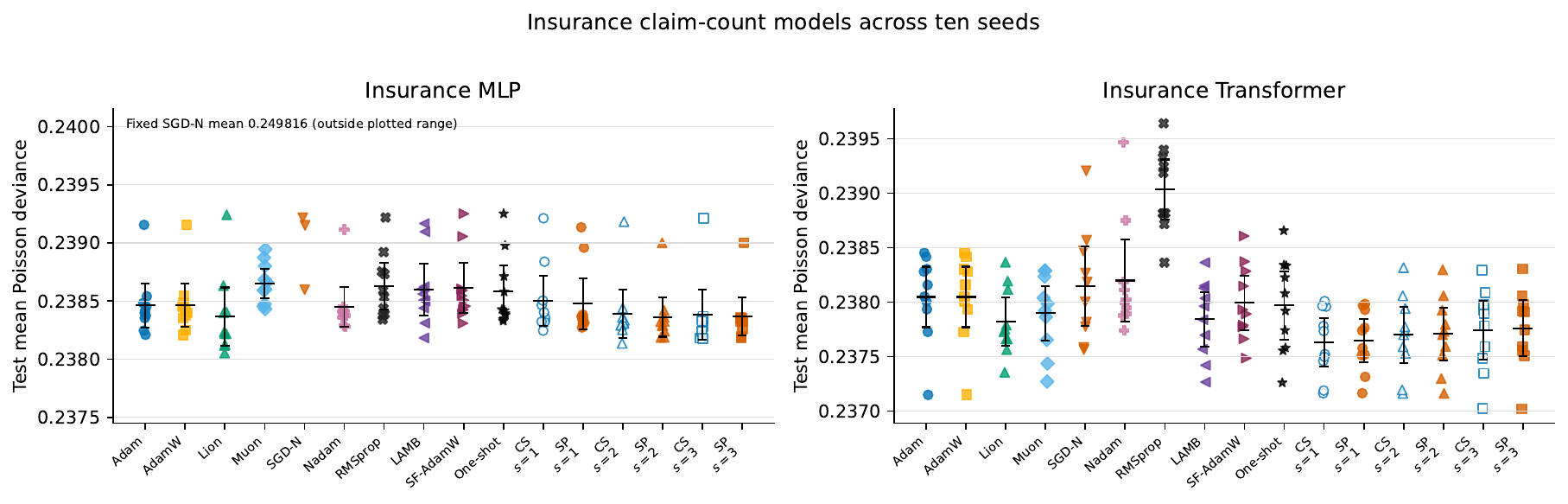}
\caption{Insurance test deviance by seed. Points are individual runs;
horizontal bars show means and 95\% confidence intervals. The MLP panel is
restricted to the range containing the competitive methods; the fixed
SGD-Nesterov mean of 0.249816 falls outside that range (smaller is better).}
\label{fig:mtpl-performance}
\end{figure*}

Tables~\ref{tab:aggregate} and \ref{tab:mtpl} report the
complete optimizer comparisons; Figures~\ref{fig:accuracy} and
\ref{fig:mtpl-performance} show the seed-level variation.

\subsubsection{Image classification}

\begin{table*}[t]
\centering
\caption{Image classification results over ten seeds. Accuracy is the mean
test accuracy in percentage points, followed by its sample standard deviation
(bigger is better).
Ranks use unrounded means across the 16 fitted training rules, with rank 1
best. The full optimizer search row is the aggregate cost of completing all
nine fixed runs and is not another fitted model.}
\label{tab:aggregate}
\footnotesize
\begin{tabular*}{\textwidth}{@{\extracolsep{\fill}}lrrrrrr@{}}
\toprule
Training rule & \multicolumn{3}{c}{MNIST} & \multicolumn{3}{c}{Fashion-MNIST} \\
\cmidrule(lr){2-4}\cmidrule(lr){5-7}
 & Accuracy & Rank & Epoch-eq. & Accuracy & Rank & Epoch-eq. \\
\midrule
Adam & 98.003$\pm$0.244 & 7 & 16.5 & 88.287$\pm$0.472 & 13 & 16.1 \\
AdamW & 97.928$\pm$0.213 & 12 & 15.0 & 88.262$\pm$0.454 & 14 & 15.8 \\
Lion & 97.552$\pm$0.189 & 16 & 13.4 & 88.578$\pm$0.299 & 9 & 12.9 \\
Muon & 98.095$\pm$0.139 & 4 & 15.2 & 88.431$\pm$0.516 & 10 & 13.3 \\
SGD-N & 97.787$\pm$0.213 & 14 & 18.1 & 88.084$\pm$0.356 & 15 & 17.6 \\
Nadam & 97.870$\pm$0.160 & 13 & 18.2 & 88.293$\pm$0.313 & 12 & 16.0 \\
RMSprop & 97.977$\pm$0.182 & 10 & 19.4 & 88.065$\pm$0.381 & 16 & 18.7 \\
LAMB & 97.690$\pm$0.309 & 15 & 21.2 & 88.419$\pm$0.406 & 11 & 26.5 \\
SF-AdamW & 98.038$\pm$0.097 & 5 & 13.7 & 89.059$\pm$0.263 & 2 & 15.5 \\
\midrule
\textbf{Full optimizer search (all 9)} & \textit{n/a} & \textit{n/a} & \textbf{150.7} & \textit{n/a} & \textit{n/a} & \textbf{152.4} \\
\midrule
One-shot & 98.116$\pm$0.119 & 3 & 20.3 & 88.839$\pm$0.442 & 8 & 22.5 \\
\midrule
CS-ROR ($s=1$) & 97.948$\pm$0.169 & 11 & 37.4 & 89.007$\pm$0.331 & 3 & 53.9 \\
SP-ROR ($s=1$) & 97.988$\pm$0.165 & 8 & 39.6 & 88.992$\pm$0.321 & 4 & 48.4 \\
CS-ROR ($s=2$) & 98.037$\pm$0.169 & 6 & 58.9 & 88.897$\pm$0.380 & 7 & 81.7 \\
SP-ROR ($s=2$) & 97.980$\pm$0.164 & 9 & 68.4 & 88.980$\pm$0.339 & 6 & 79.8 \\
CS-ROR ($s=3$) & \textbf{98.132$\pm$0.182} & \textbf{1} & 116.1 & 88.988$\pm$0.359 & 5 & 110.7 \\
SP-ROR ($s=3$) & 98.131$\pm$0.099 & 2 & 110.7 & \textbf{89.068$\pm$0.338} & \textbf{1} & 126.9 \\
\bottomrule
\end{tabular*}

\end{table*}

On MNIST, fixed Muon reaches \(98.095\%\). One-epoch CS-ROR and SP-ROR reach
\(97.948\%\) and \(97.988\%\), using 37.4 and 39.6 epoch-equivalents rather
than 150.7 for the full optimizer search. Extending the scout to three epochs
raises the ROR means to \(98.132\%\) and \(98.131\%\), but roughly triples the
search cost. The paired \(s=1\) minus \(s=3\) difference is
\(-0.184\) percentage points for CS-ROR, with a 95\% confidence interval of
\([-0.355,-0.013]\), and \(-0.143\) points for SP-ROR, with interval
\([-0.307,0.021]\). MNIST therefore shows the main trade-off: the shortest
scout is much cheaper, with a small loss of accuracy.

On Fashion-MNIST, the ROR means lie between \(88.897\%\) and \(89.068\%\);
fixed ScheduleFreeAdamW reaches \(89.059\%\). One-epoch CS-ROR and SP-ROR cost
53.9 and 48.4 epoch-equivalents, compared with 152.4 for the full search.
Every paired comparison between a shorter scout and \(s=3\) includes zero, so
the experiment provides no clear evidence that longer scouting improves
Fashion-MNIST accuracy.

\newpage
\subsubsection{Insurance claim counts}

\begin{table*}[t]
\centering
\caption{Insurance results over ten seeds. Test deviance is the mean Poisson
deviance, followed by its sample standard deviation (smaller is better). Calibration is predicted
claims divided by observed claims. Ranks use unrounded mean deviance across the
16 fitted training rules within each architecture, with rank 1 best. The full
optimizer search row is the aggregate cost of completing all nine fixed runs
and is not another fitted model.}
\label{tab:mtpl}
\footnotesize
\begin{tabular*}{\textwidth}{@{\extracolsep{\fill}}llrcrr@{}}
\toprule
Architecture & Training rule & Test deviance & Rank & Calibration & Epoch-eq. \\
\midrule
MLP & Adam & 0.238463$\pm$0.000263 & 7 & 0.999 & 44.3 \\
 & AdamW & 0.238463$\pm$0.000263 & 8 & 0.999 & 44.2 \\
 & Lion & 0.238365$\pm$0.000353 & 2 & 0.995 & 31.3 \\
 & Muon & 0.238652$\pm$0.000178 & 15 & 0.999 & 22.1 \\
 & SGD-N & 0.249816$\pm$0.007473 & 16 & 1.003 & 58.1 \\
 & Nadam & 0.238449$\pm$0.000240 & 6 & 1.000 & 47.4 \\
 & RMSprop & 0.238629$\pm$0.000282 & 14 & 0.997 & 91.5 \\
 & LAMB & 0.238599$\pm$0.000311 & 12 & 0.998 & 71.0 \\
 & SF-AdamW & 0.238614$\pm$0.000302 & 13 & 1.003 & 32.5 \\
\midrule
 & \textbf{Full optimizer search (all 9)} & \textit{n/a} & \textit{n/a} & \textit{n/a} & \textbf{442.4} \\
\midrule
 & One-shot & 0.238581$\pm$0.000311 & 11 & 1.001 & 44.8 \\
\midrule
 & CS-ROR ($s=1$) & 0.238500$\pm$0.000300 & 10 & 0.999 & 107.8 \\
 & SP-ROR ($s=1$) & 0.238476$\pm$0.000305 & 9 & 1.001 & 110.0 \\
 & CS-ROR ($s=2$) & 0.238391$\pm$0.000289 & 5 & 0.999 & 209.0 \\
 & SP-ROR ($s=2$) & \textbf{0.238362$\pm$0.000236} & \textbf{1} & 0.999 & 199.5 \\
 & CS-ROR ($s=3$) & 0.238382$\pm$0.000300 & 4 & 0.999 & 280.8 \\
 & SP-ROR ($s=3$) & 0.238368$\pm$0.000232 & 3 & 0.999 & 256.5 \\
\midrule
\midrule
CLS Transformer & Adam & 0.238047$\pm$0.000388 & 13 & 1.021 & 11.2 \\
 & AdamW & 0.238047$\pm$0.000388 & 12 & 1.021 & 11.2 \\
 & Lion & 0.237817$\pm$0.000311 & 7 & 1.007 & 11.9 \\
 & Muon & 0.237897$\pm$0.000352 & 9 & 1.006 & 13.0 \\
 & SGD-N & 0.238143$\pm$0.000510 & 14 & 1.027 & 23.3 \\
 & Nadam & 0.238198$\pm$0.000523 & 15 & 1.024 & 13.3 \\
 & RMSprop & 0.239031$\pm$0.000388 & 16 & 0.980 & 19.8 \\
 & LAMB & 0.237842$\pm$0.000351 & 8 & 1.020 & 23.3 \\
 & SF-AdamW & 0.237993$\pm$0.000350 & 11 & 1.014 & 13.8 \\
\midrule
 & \textbf{Full optimizer search (all 9)} & \textit{n/a} & \textit{n/a} & \textit{n/a} & \textbf{140.8} \\
\midrule
 & One-shot & 0.237968$\pm$0.000435 & 10 & 1.015 & 18.9 \\
\midrule
 & CS-ROR ($s=1$) & \textbf{0.237627$\pm$0.000311} & \textbf{1} & 1.007 & 42.9 \\
 & SP-ROR ($s=1$) & 0.237646$\pm$0.000279 & 2 & 1.007 & 40.7 \\
 & CS-ROR ($s=2$) & 0.237699$\pm$0.000359 & 3 & 1.011 & 66.5 \\
 & SP-ROR ($s=2$) & 0.237707$\pm$0.000340 & 4 & 1.012 & 64.6 \\
 & CS-ROR ($s=3$) & 0.237740$\pm$0.000380 & 5 & 1.009 & 99.9 \\
 & SP-ROR ($s=3$) & 0.237758$\pm$0.000360 & 6 & 1.012 & 94.5 \\
\bottomrule
\end{tabular*}

\end{table*}

In Table~\ref{tab:mtpl}, Adam and AdamW agree to the displayed precision on
both architectures. This is a rounding coincidence rather than a duplicated
run.\footnote{The unrounded means differ, \(0.23846287\) for Adam versus
\(0.23846336\) for AdamW on the MLP and \(0.23804675\) versus \(0.23804673\)
on the Transformer, which is why the two rows carry different ranks. Keras
applies decoupled weight decay scaled by the learning rate, so decay
\(10^{-4}\) at learning rate \(10^{-3}\) shrinks the weights by a factor of
\(1-10^{-7}\) per update. Over the few thousand updates of these small
networks, that effect is below the displayed precision; the image tasks in
Table~\ref{tab:aggregate} show a visible Adam--AdamW gap.}

On the MLP, fixed Lion reaches a mean deviance of \(0.238365\). One-epoch
CS-ROR and SP-ROR reach \(0.238500\) and \(0.238476\), using 107.8 and 110.0
epoch-equivalents rather than 442.4 for the full optimizer search. Two-epoch
SP-ROR has the lowest observed loss, \(0.238362\), at a cost of 199.5. The
paired comparisons with \(s=3\) all include zero, including the \(s=1\)
comparisons, so there is no clear evidence that the longer scout improves MLP
deviance.

Fixed Muon ranks 15 of 16 on the MLP, which stands in contrast to the tabular
benchmark of Gorishniy et al. \cite{gorishniy2026optimizers}, where Muon
consistently outperformed AdamW. Three differences plausibly contribute: our
learning rate of \(10^{-3}\) is fixed in advance rather than tuned for this
task, the 4,791-parameter network is far smaller than typical tabular
benchmark models, and the embedding-heavy parameterization routes most
parameters to Muon's AdamW branch rather than to its orthogonalized update. Finally, a Poisson deviance might favour different optimizers than the loss functions used in \cite{gorishniy2026optimizers}.

On the Transformer, fixed Lion reaches \(0.237817\) (smaller is better). One-epoch CS-ROR and
SP-ROR reach \(0.237627\) and \(0.237646\), at costs of 42.9 and 40.7
epoch-equivalents rather than 140.8 for the full search. These are the lowest
observed losses in the table, but their paired differences from Lion include
zero. All shorter-scout comparisons with \(s=3\) also include zero.

\begin{figure*}[t]
\centering
\includegraphics[width=0.94\textwidth]{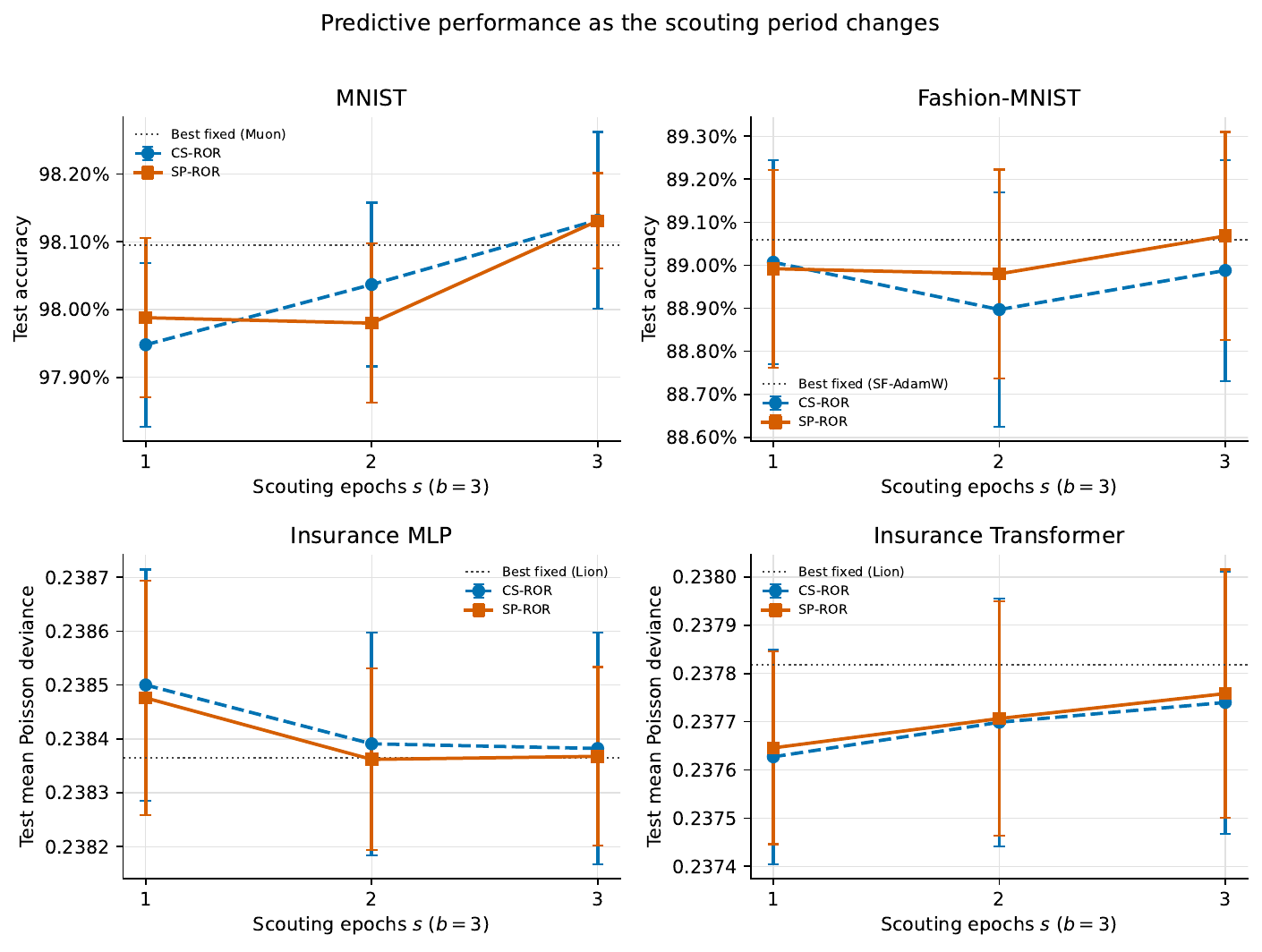}
\caption{Predictive performance as the scouting period changes from one to
three epochs, with retained segment length fixed at \(b=3\).
Classification accuracy is higher when better; Poisson deviance is lower when
better. Error bars show 95\% confidence intervals for the mean.}
\label{fig:scout-length}
\end{figure*}

Figure~\ref{fig:scout-length} shows the same pattern across tasks. Increasing
\(s\) raises cost sharply because all nine candidates receive the extra scout
epochs, while predictive performance changes much less. Naturally, the first
gradient descent steps are the most significant ones and losses flatten after
a few epochs. That is why short scouting is good enough. The paired
short-scout comparisons are reported in Appendix~\ref{app:paired-comparisons}.

\FloatBarrier
\clearpage
\onecolumn
\section{Interpreting the schedules}

The schedules show which optimizer produced each accepted segment. In SP-ROR,
a repeated label means that the incumbent was selected again and resumed with
its saved state. A changed label means that ROR switched to a challenger whose
state was initialized at the start of that tournament.

\begin{figure}[h]
\centering
\includegraphics[width=0.88\textwidth]{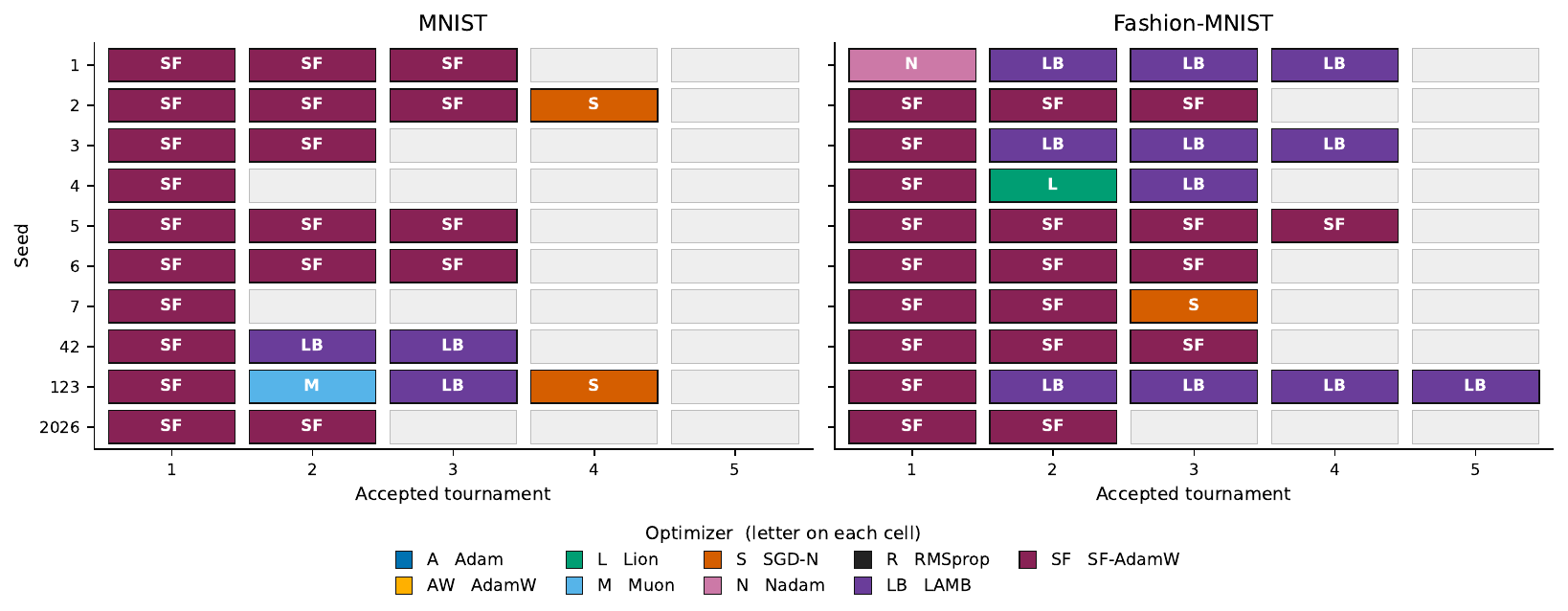}
\par\vspace{2pt}
\includegraphics[width=0.88\textwidth]{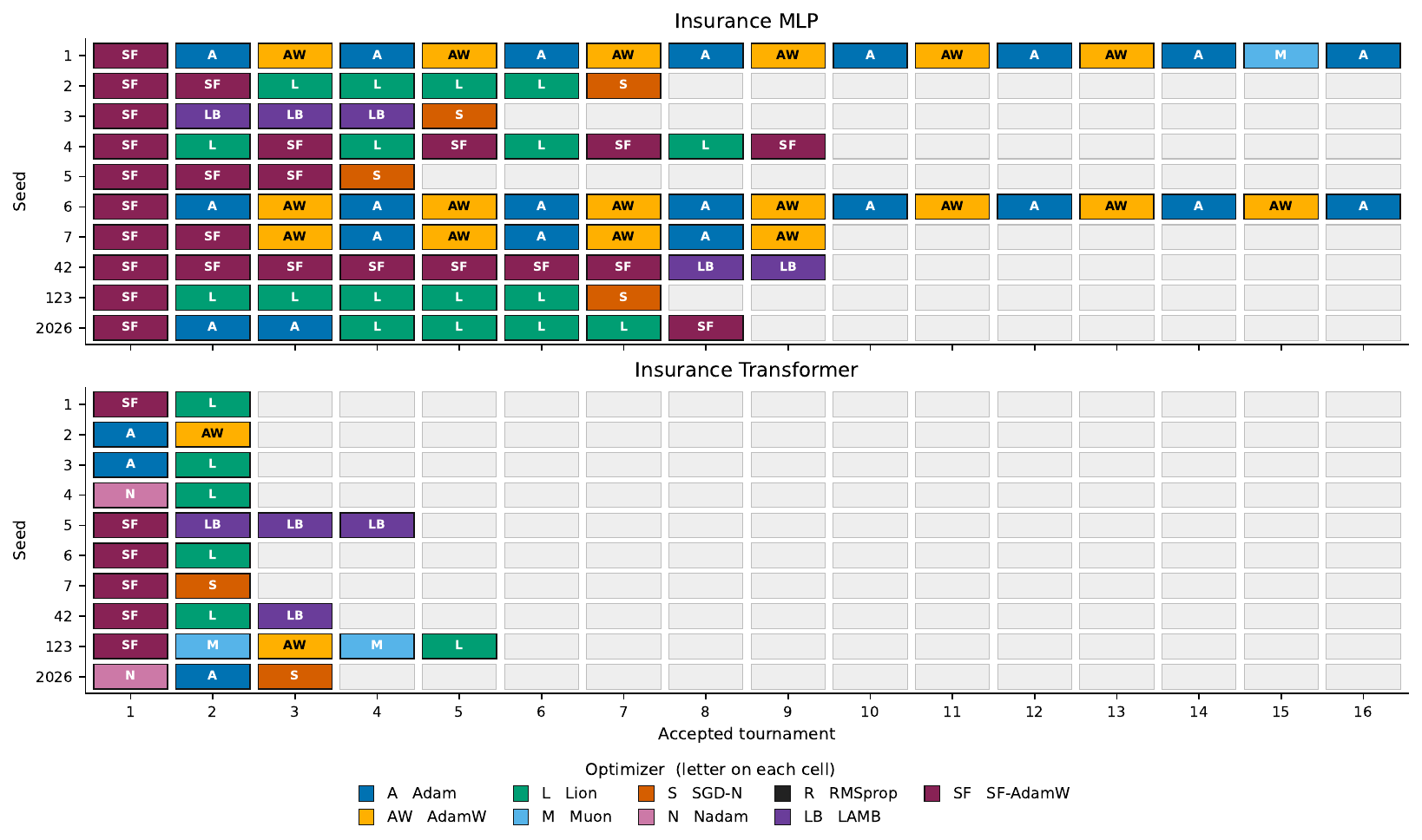}
\caption{State-preserving ROR schedules ($s=1$, $b=3$). Image tasks above,
insurance models below. Blank cells follow the first rejected tournament.
Letters in each cell are defined in the legend.}
\label{fig:schedules}
\end{figure}

The image schedules are comparatively stable
(Figure~\ref{fig:schedules}). MNIST averages 2.6 accepted tournaments and 0.5
switches per seed. All ten paths start with ScheduleFreeAdamW; seven also end
there. Fashion-MNIST averages 3.4 accepted tournaments and 0.6 switches. Nine
paths start with ScheduleFreeAdamW, and the final choices are split mainly
between ScheduleFreeAdamW and LAMB.

The Insurance MLP changes optimizer more often. Its paths average 9.0 accepted
tournaments and 5.6 switches. All start with ScheduleFreeAdamW, but the final
optimizers are spread across SGD-Nesterov, Adam, ScheduleFreeAdamW, AdamW, and
LAMB. The Insurance Transformer averages 2.7 accepted tournaments and 1.5 switches, and
half of its paths finish with Lion.

These schedules support the motivation for checking optimizer choice more
than once: the optimizer favored after one epoch is not always the optimizer
favored later. They also show that ROR does not switch mechanically at every
tournament. On the image tasks it usually retains the incumbent, whereas the
Insurance MLP makes frequent changes. The corresponding cold-start schedules,
where a repeated label does not carry state forward, are reported in
Appendix~\ref{app:cold-schedules}.

\clearpage
\twocolumn
\section{Caveats}

The experiments cover four tasks, ten seeds, and relatively small networks.
They test the optimizer and learning-rate pairs listed in
Section~3, not every possible tuning of each optimizer family. A different
learning rate could change both the fixed-optimizer ranking and the ROR
schedule.

SP-ROR preserves only the incumbent optimizer's state. Challengers start with
new state, so a tournament compares continuation with fresh alternatives. It
is not possible to give every challenger its own exact history on the retained
weights. Doing so would require maintaining nine matching model trajectories,
which is the parallel search that ROR is meant to avoid.

The same validation set is used at every tournament. Repeated queries may
eventually favor a schedule that fits that validation sample. A larger search
would need a reusable holdout, nested validation, or cross-validation. The
one-shot baseline also resets the chosen optimizer before continuation, so it
should be read as a cheap practical alternative rather than a state-matched
version of ROR.

Epoch-equivalents count all training epochs, including discarded scouts, but
they treat every epoch as equally expensive. This ignores differences in
per-update work, including Muon's matrix operations. Hardware-normalized time,
energy, or floating-point operations would provide a fuller cost comparison.
The fixed data order and first-rejection stopping rule also make these
experiments less noisy than training with data augmentation or reshuffling.
Such settings would probably need a tournament-level patience rule. The
stopping rules used here are also somewhat asymmetric: fixed baselines receive five epochs of
patience before stopping, whereas ROR stops at its first rejected tournament,
an effective patience of a single \(b\)-epoch segment. This asymmetry can end
an ROR run that a later tournament would have improved, so it works against
ROR's predictive results while flattering its epoch-equivalent costs.

\FloatBarrier
\section{Conclusion}

We asked whether optimizer selection could happen during training instead of
before it. ROR answers that question with a sequence of short tournaments.
Every optimizer scouts from the current weights, the best scout completes the
next segment, and only that segment is retained. This creates one model path
whose optimizer may change as training progresses.

The experiments show that the length of the scout matters more for cost than
for prediction. A plausible explanation is that much of the reduction in validation loss occurs during the earliest optimization steps, after which losses often flatten. This may explain why short scouting performs well. With \(b=3\), reducing \(s\) from three epochs to one lowers ROR cost by 51\% to 68\%. One-epoch ROR uses only 24\% to 35\% of the aggregate training required to complete all nine fixed-optimizer runs. It
finishes close to the best fixed optimizer on all four tasks, although MNIST
shows a small accuracy cost for the shorter scout.

ROR did not show that an optimizer schedule chosen by ROR necessarily predicts better than the best
fixed optimizer. Some observed means favor ROR, especially on the Insurance
Transformer, but every paired comparison with the best fixed optimizer
includes zero. The schedules do show that the preferred optimizer can change
within a run, particularly on the Insurance MLP. A larger study is needed to
decide whether those changes can reliably outperform a well-tuned fixed
choice.

The one-shot baseline sharpens this reading. It is cheaper than one-epoch ROR
on every task, beats it on MNIST, and trails its observed means on the other
three tasks. When the optimizer ranking after one epoch persists, a
single early comparison can capture most of the value of optimizer selection.
The case for repeated tournaments - as we do in ROR - rests on tasks where the preferred
optimizer changes during training; the price being roughly a doubling of the one-shot cost.

Preserving the incumbent's optimizer state is conceptually cleaner when the
same optimizer wins again, but it has no consistent predictive or
computational advantage here. The practical lesson is simpler: if the
optimizer is not known in advance, one-epoch scouting provides most of ROR's
benefit at a fraction of the cost of exhaustive search. Longer scouts buy more
evidence at each tournament, but they should be justified by a task where that
extra evidence improves the final model.

\section*{Use of generative AI}

OpenAI Codex assisted with software development, experiment orchestration,
language editing, and the preparation of tables and figures. The authors
designed the study, specified the methods and comparisons, reviewed the
implementation and results, interpreted the findings, and take responsibility
for the final manuscript.

\clearpage
\onecolumn
\appendix

\section{Paired short-scout comparisons}
\label{app:paired-comparisons}

Table~\ref{tab:paired-short-scout} compares \(s=1\) and \(s=2\), i.e. scouting choices less than $b$, directly with
the \(s=3\) design, keeping \(b=3\) and matching seeds. Negative
accuracy differences favor \(s=3\); negative deviance differences favor the
shorter scout. Fourteen of the sixteen confidence intervals include zero. The
exceptions are the MNIST comparisons for CS-ROR at \(s=1\) and SP-ROR at
\(s=2\), both of which favor \(s=3\). No adjustment is made for multiple
comparisons.

\begin{table}[h]
\centering
\caption{Paired short-scout comparisons over ten seeds. Differences are the
shorter scout minus \(s=3\). Search cost is reported in epoch-equivalents.}
\label{tab:paired-short-scout}
\footnotesize
\begin{tabular*}{\textwidth}{@{\extracolsep{\fill}}lllrcrr@{}}
\toprule
Task & Policy & Comparison & Metric & Difference (95\% CI) & Short eq. & $s=3$ eq. \\
\midrule
MNIST & CS-ROR & $s=1$ minus $s=3$ & Accuracy (pp) & -0.184 [-0.355, -0.013] & 37.4 & 116.1 \\
 & CS-ROR & $s=2$ minus $s=3$ & Accuracy (pp) & -0.095 [-0.256, 0.066] & 58.9 & 116.1 \\
 & SP-ROR & $s=1$ minus $s=3$ & Accuracy (pp) & -0.143 [-0.307, 0.021] & 39.6 & 110.7 \\
 & SP-ROR & $s=2$ minus $s=3$ & Accuracy (pp) & -0.151 [-0.285, -0.017] & 68.4 & 110.7 \\
\addlinespace[2pt]
Fashion-MNIST & CS-ROR & $s=1$ minus $s=3$ & Accuracy (pp) & 0.019 [-0.258, 0.296] & 53.9 & 110.7 \\
 & CS-ROR & $s=2$ minus $s=3$ & Accuracy (pp) & -0.091 [-0.377, 0.195] & 81.7 & 110.7 \\
 & SP-ROR & $s=1$ minus $s=3$ & Accuracy (pp) & -0.076 [-0.302, 0.150] & 48.4 & 126.9 \\
 & SP-ROR & $s=2$ minus $s=3$ & Accuracy (pp) & -0.088 [-0.310, 0.134] & 79.8 & 126.9 \\
\addlinespace[2pt]
Insurance MLP & CS-ROR & $s=1$ minus $s=3$ & Poisson deviance & 0.000118 [-0.000031, 0.000266] & 107.8 & 280.8 \\
 & CS-ROR & $s=2$ minus $s=3$ & Poisson deviance & 0.000009 [-0.000036, 0.000053] & 209.0 & 280.8 \\
 & SP-ROR & $s=1$ minus $s=3$ & Poisson deviance & 0.000108 [-0.000035, 0.000252] & 110.0 & 256.5 \\
 & SP-ROR & $s=2$ minus $s=3$ & Poisson deviance & -0.000006 [-0.000042, 0.000031] & 199.5 & 256.5 \\
\addlinespace[2pt]
Insurance Transformer & CS-ROR & $s=1$ minus $s=3$ & Poisson deviance & -0.000113 [-0.000266, 0.000041] & 42.9 & 99.9 \\
 & CS-ROR & $s=2$ minus $s=3$ & Poisson deviance & -0.000041 [-0.000170, 0.000087] & 66.5 & 99.9 \\
 & SP-ROR & $s=1$ minus $s=3$ & Poisson deviance & -0.000113 [-0.000239, 0.000014] & 40.7 & 94.5 \\
 & SP-ROR & $s=2$ minus $s=3$ & Poisson deviance & -0.000052 [-0.000133, 0.000030] & 64.6 & 94.5 \\
\bottomrule
\end{tabular*}

\end{table}

\FloatBarrier

\section{Cold-start schedules}
\label{app:cold-schedules}

In CS-ROR, every scout starts with newly initialized optimizer state. A
repeated label therefore means that the same optimizer type was selected
again, not that its momentum or adaptive moments continued.

With \(s=1\) and \(b=3\), CS-ROR averages 2.4 accepted tournaments and 0.8
switches on MNIST, 3.9 and 1.1 on Fashion-MNIST, 8.8 and 2.3 on the Insurance
MLP, and 2.9 and 1.5 on the Insurance Transformer. The Insurance MLP has 55 repeated
labels across the ten seeds. These are repeated cold starts, which explains
why a schedule can look stable even though optimizer state is discarded at
each tournament.

\begin{figure}[h]
\centering
\includegraphics[width=\textwidth]{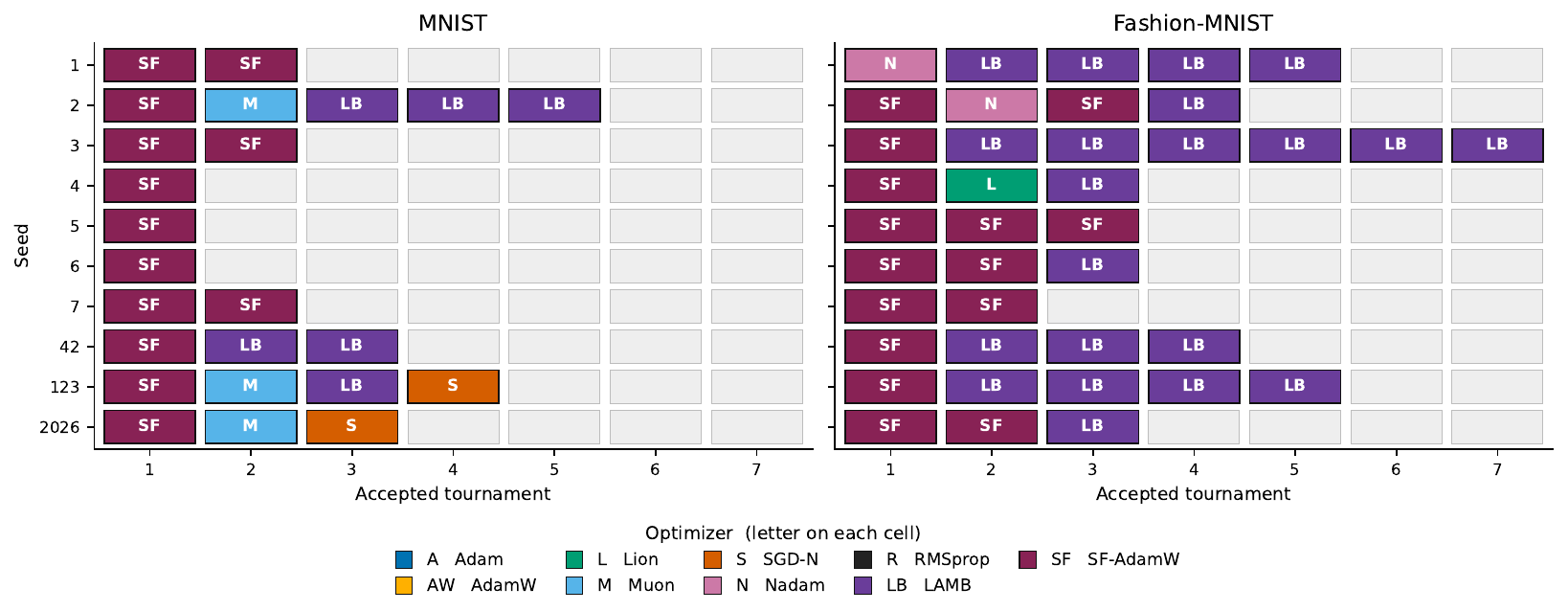}
\par\vspace{6pt}
\includegraphics[width=\textwidth]{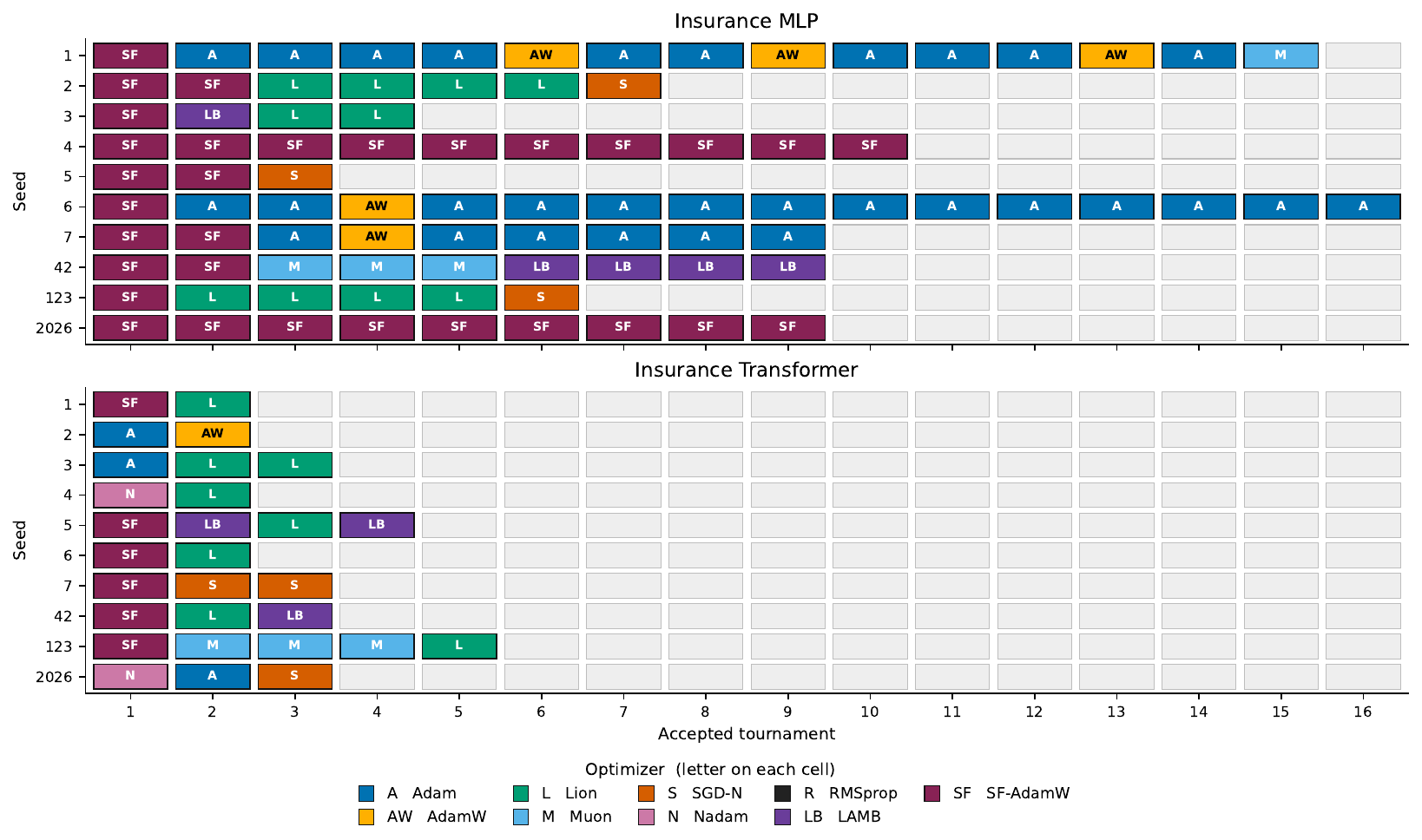}
\caption{Cold-start ROR schedules ($s=1$, $b=3$). Image tasks above,
insurance models below. Every scout, including a repeated winner, starts with
newly initialized optimizer state. Letters in each cell are defined in the legend.}
\label{fig:schedules-cold}
\end{figure}

\FloatBarrier
\clearpage

\section{Keras 3 pseudocode}
\label{app:pseudocode}
\thispagestyle{plain}

With \texttt{preserve\_state=True}, a repeated incumbent restores its state;
with \texttt{False}, every scout starts fresh. In both cases the selected scout
keeps its state for the remaining
\texttt{segment\_epochs - scout\_epochs} epochs.

\begin{center}
\begin{minipage}{0.90\textwidth}
\begin{lstlisting}[style=kerascode,caption={SP-ROR and CS-ROR in pure Keras 3 pseudocode.},label={lst:keras}]
import keras

optimizer_fns = {
    "adam": make_adam,
    # Seven other optimizer constructors are omitted here.
    "schedule_free_adamw": make_sf_adamw,
}

def train(weights, make_optimizer, epochs, state=None):
    model = make_model()
    model.set_weights(weights)
    optimizer = make_optimizer()
    model.compile(optimizer=optimizer, loss=loss_fn)
    if state is not None:
        optimizer.build(model.trainable_variables)
        for variable, value in zip(optimizer.variables, state):
            variable.assign(value)
    if epochs:
        model.fit(
            x_train, y_train,
            validation_data=(x_valid, y_valid),
            epochs=epochs, shuffle=False, verbose=0,
        )
    score = validation_objective(model, x_valid, y_valid)
    new_state = [keras.ops.convert_to_numpy(v)
                 for v in optimizer.variables]
    return model.get_weights(), new_state, score

weights = make_model().get_weights()
incumbent = evaluate_weights(weights, valid_data)
incumbent_name = None
incumbent_state = None
preserve_state = True
scout_epochs = 1
segment_epochs = 3

while True:
    scouts = [
        (name, *train(
            weights, fn, scout_epochs,
            state=incumbent_state
            if preserve_state and name == incumbent_name
            else None,
        ))
        for name, fn in optimizer_fns.items()
    ]
    name, scout_weights, scout_state, scout_score = min(
        scouts, key=lambda candidate: candidate[3]
    )

    if scout_epochs < segment_epochs:
        new_weights, new_state, new_score = train(
            scout_weights,
            optimizer_fns[name],
            segment_epochs - scout_epochs,
            state=scout_state,
        )
    else:
        new_weights = scout_weights
        new_state = scout_state
        new_score = scout_score

    if incumbent - new_score <= min_delta:
        break
    weights = new_weights
    incumbent_name = name
    incumbent_state = new_state
    incumbent = new_score
\end{lstlisting}
\end{minipage}
\end{center}

\end{document}